\documentclass{INTERSPEECH2023}
\usepackage{float}                  
\usepackage{subfig}                 
\usepackage{overpic}                
\usepackage{multirow}
\usepackage{makecell}

\interspeechcameraready

\title{DistilXLSR: A Light Weight Cross-Lingual Speech Representation Model}
\name{Haoyu Wang$^1$, Siyuan Wang$^1$, Wei-Qiang Zhang$^1$\sthanks{* Corresponding author}, Jinfeng Bai$^2$\thanks{This work was supported by the National Key R\&D Program of China under Grant No. 2020AAA0104500, and the National Natural Science Foundation of China under Grant No. 62276153.}}
\address{
  $^1$Department of Electronic Engineering, Tsinghua University, Beijing 100084, China \\
  $^2$TAL Education, Beijing 100084, China
}
\email{w-hy21@mails.tsinghua.edu.cn, wq-zhang@tsinghua.edu.cn}

\begin{document}

\maketitle
 
\begin{abstract}
Multilingual self-supervised speech representation models have greatly enhanced the speech recognition performance for low-resource languages, and the compression of these huge models has also become a crucial prerequisite for their industrial application. In this paper, we propose DistilXLSR, a distilled cross-lingual speech representation model. By randomly shuffling the phonemes of existing speech, we reduce the linguistic information and distill cross-lingual models using only English data. We also design a layer-jumping initialization method to fully leverage the teacher's pre-trained weights. Experiments on 2 kinds of teacher models and 15 low-resource languages show that our method can reduce the parameters by 50\% while maintaining cross-lingual representation ability. Our method is proven to be generalizable to various languages/teacher models and has the potential to improve the cross-lingual performance of the English pre-trained models.
\end{abstract}

\section{Introduction}

Self-supervised pre-trained models have made many significant breakthroughs in low-resource speech recognition. By learning from a large amount of multilingual unlabeled data, these self-supervised pre-trained models can provide cross-lingual phoneme-level representations for almost any language. Models fine-tuned from multilingual pre-trained models can achieve satisfactory word error rates (WER) with extremely limited or even no speech data \cite{zhao_improving_2022, yi_applying_2021,xu_simple_2021, gao_zero-shot_2021}.

However, these multilingual pre-trained models, represented by XLS-R and XLSR53, typically have hundreds of millions of parameters, which is an obstacle to their application on mobile devices such as laptops and smartphones. Considering the excellent performance of these models in low-resource speech recognition, a compressed multilingual speech representation model is of undoubted importance to the industrial application of speech recognition in minority languages.

Model pruning is an efficient method to reduce the parameters of the pre-trained models. The lottery ticket hypothesis assumes that a sparse subnetwork can be extracted from a dense network without sacrificing the performance \cite{frankle_lottery_2019, chen_lottery_2020}. PARP proposes a feasible measure to discover the subnetwork from self-supervised speech representation models by alternating pruning and fine-tuning \cite{lai_parp_2021}. Similarly, by alternate quantization and pruning, Wang et al. successfully remove 50\% of the parameters from the Wav2vec 2.0 model and quantize it down to a 4-bit precision \cite{wang_structured_2020}. Although these pruning-based compression methods retain most of the performance, their acceleration still requires the support of specific hardware devices.

Knowledge distillation is a hardware-friendly way to transfer the representation ability to a compact student that can be used in normal computing devices. DistilHuBERT compresses a 12-layer Hubert-based model to get a 2-layer student model and appreciably reduces the model size \cite{chang_distilhubert_2022}.  FitHuBERT designs a thin but deep student model to improve the representation ability of the student model and achieves better performance with fewer parameters than the DistilHuBERT model \cite{lee_fithubert_2022}. 

Compared to the Hubert base model used in previous studies, the distillation of cross-lingual speech representation models faces new challenges. First, it is difficult to obtain training data for low-resource languages, collecting and formatting data from multiple languages also requires time and effort. To address this challenge, we found inspiration in the
RNN-transducer (RNN-T) domain adaptation problem. Zhao et al. proposed a data splicing method, which randomly selects speech segments from existing data to generate new training utterances \cite{zhao_addressing_2021}. This method can adapt a pre-trained RNN-T model to new domains with negligible cost. For different languages, the phonotactics of a sentence is one of the most important features \cite{zissman_automatic_1994}. Therefore, we want to distill the multilingual pre-trained models using only unlabeled English data by randomly selecting phonemes from existing utterances.

Second, the parameters of large pre-trained models such as XLSR-53 have more complex interrelationships, which is a barrier for the learning of the student. Therefore, we design a layer-jumping initialization method to better exploit the pre-trained parameters and retain the inter-layer similarity of the teacher.

 In this paper, we propose DistilXLSR, a compact multilingual speech representation model\footnote{Aavailable at https://github.com/backspacetg/distilXLSR}. We verify the effectiveness of our method on XLS-R and XLSR-53. Experiments on 15 low-resource languages prove that our method can maintain most of the performance and achieves comparable performance with multilingual distillation.

\section{Method}
\subsection{Wav2vec 2.0 Models}
XLS-R and XLSR-53 are two of the most commonly used multilingual pre-trained models, and both can be considered as multilingual versions of the Wav2vec 2.0 model. Wav2vec 2.0 models are composed of a CNN feature extractor and a multi-layer transformer encoder \cite{baevski_wav2vec_2020}. For the XLS-R and XLSR-53 models, the feature extractors have 6 CNN layers and the encoders have 24 layers. These models are trained through the contrastive prediction coding (CPC) task, where the future frames ought to be distinguished from some randomly sampled distractors. Through CPC training, the outputs of each transformer layer, or the hidden states, will contain higher-level information about the input audio. 

Our distilled model also has a similar structure to the XLS-R and the XLSR-53 models. Considering that an excessively large difference in size may have a negative effect on distillation \cite{lu_knowledge_2022}, we decide to use a 12-layer transformer encoder which leads to around a 50\% reduction in the number of parameters.
\subsection{Distillation Objective}
Typically, in knowledge distillation, the student model tries to learn from the teacher model by mimicking the teacher's behavior. For transformer-based wav2vec 2.0 teachers, the students usually learn from the hidden states, the attention score, or the logits of the CPC task. Some previous works \cite{lu_knowledge_2022,nouriborji_minialbert_2022} and our preliminary experiments show that minimizing the mean square error (MSE) of the hidden states and using multi-task distillation to learn from different depths will lead the students to the best performance. Formally speaking, let $H$ be the hidden states of the teacher model and $\hat{H}$ be those of the student model, the distillation loss is computed as follows:

\vspace{-1.2em}
\begin{align}
l_\text{distil}(\hat{H}, H) = \sum_{(i,j)\in S}l_\text{MSE}(h_i, \hat{h}_j)
\label{distil}
\end{align}

Each tuple $(i, j)$ in $S$ denotes a student-teacher layer pair where $h_i$, $\hat{h}_j$ are the hidden states from layer $i$ and $j$ of the student and teacher model, respectively. 

\vspace{-0.3em}
\subsection{Layer-Jumping Initialization}

Speech signal contains a lot of information. Emotion, prosody, and semantic information can all be encoded in an utterance. In pre-trained speech representation models, it is widely regarded that hidden states from different layers contain different kinds of information \cite{chang_distilhubert_2022,chen_optimize_2022}. Phoneme-level semantic information is usually contained in the last few layers. As a result, learning from these layers helps to achieve better performance in speech recognition tasks.

In previous works, the student models usually load weights from the lower teacher layers (e.g., the first two transformer layers, depending on the number of layers of the student model) or are simply trained from scratch \cite{chang_distilhubert_2022}. However, we assume that this may not be appropriate for larger pre-trained models such as XLS-R or XLSR-53. Fig \ref{base} and \ref{xlsr} show the Centered Kernel Alignment (CKA) inter-layer similarity \cite{kornblith_similarity_2019} of the wav2vec 2.0 base model and the XLSR-53 model, respectively. The CKA similarity, which is based on the inner product of the hidden states, shows that the last few layers of the XLSR-53 model are more different from the previous ones, and the relationship between these layers is more complex.

It makes intuitive sense that directly loading the weight of the last few layers would help deal with such complexity. As a result, we propose the layer-jumping initialization method, where the teacher layers are selected at intervals when initializing the student, to take full advantage of the pre-trained parameters. Formally speaking, the student layer $\hat{h_s^i}$ is initialized as
\begin{align}
    \hat{\theta}_s^i=\theta_t^{2i}\text{,}
\end{align}where $\hat{\theta_s^i}$ are the parameters of student layer $i$ and $\theta_t^{2i}$ are those of teacher layer $2i$. Due to the layer drop strategy, where the transformer layers are randomly dropped during pre-training, so the teacher models are actually robust to such deletion of layers.

\subsection{Data Splicing}
Using only English data to distill cross-lingual pre-trained models can help to fully utilize the large English datasets. Moreover, pre-training can also benefit from similar techniques and the cross-lingual representation ability of English pre-trained models can be improved.

To reduce the language-dependent information in an English speech utterance, we randomly shuffle the syllables in the utterances. The reason for choosing syllables rather than phonemes as the basic unit in our data splicing method is to keep the phoneme context coherent and to avoid continuous multiple constants which are rare in human languages.

\begin{figure}[t]
    \centering
    \includegraphics[width=\linewidth]{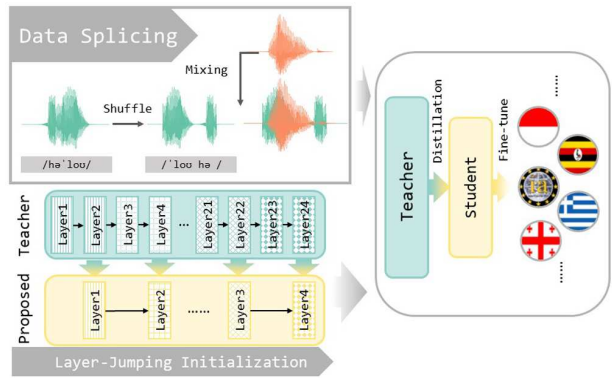}
    \caption{An overview of our method. Syllables in existing utterances are shuffled to get training data with less language-dependent information.}
    \label{method}
    \vspace{-2em}
\end{figure}

We train a Gaussian Mixture Model - Hidden Markov Model (HMM-GMM) to align the audio with the phoneme sequences. We add syllable-separating symbols to all the pronunciations in the lexicon and tag the utterances with syllable-level timestamps. During training, the syllables in an utterance are shuffled and spliced into a new speech with less language-dependent information. Fig \ref{method} provides an overview of our method.

\section{Experiments}

\textbf{Datasets}. For distillation, we only use the Librispeech English dataset \cite{panayotov_librispeech_2015}; for fine-tuning, we select 15 languages from the MATERIAL\footnote{https://www.iarpa.gov/index.php/research-programs/material}, Babel \cite{gales_speech_nodate} and Common Voice datasets \cite{ardila_common_2020}. 

Table \ref{tab:data} shows the details of our datasets. For the languages from MATERIAL and Babel, the datasets are provided by the OpenASR21 challenge\footnote{https://sat.nist.gov/openasr21}, which is a track of the NIST Open Speech Analytic Technologies (OpenSAT) evaluations. A 10-hour training set and a 10-hour development set are provided for each of the languages, consisting mainly of telephone conversations. 10 languages from the MATERIAL and Babel datasets are used for fine-tuning. To compare the result
between data splicing and real-world multilingual distillation, we also select 5 additional languages to ensure that the multilingual distillation and fine-tuning sets do not overlap. 

Common Voice is a cloud-sourced multilingual dataset with clearer speech quality. Considering that the MATERIAL and Babel datasets contain mainly African and Asian languages, we select 5 European languages from the Common Voice dataset, and randomly sample a 5-hour subset to simulate a low-resource scenario for each language. All the training audio is resampled to 16KHz.

\begin{table}[t]
    \caption{The low-resource languages. Besides the fine-tuning languages, we also select 5 languages to compare the result between data splicing and real-world multilingual distillation.}
    \centering
    \begin{tabular}{lll}
        \toprule
        \textbf{Split} & \textbf{Source} & \textbf{Languages} \\
        \midrule
        \multirow{8}{*}{Fine-tune} & MATERIAL & Tamil (ta), Farsi (fa) \\
        \cline{2-3}
         & \makecell[l]{Common\\ Voice}  & \makecell[l]{Basque (eu), Dutch (nl), \\Greek (el),\\ Interlingua (ia), Polish (pl)\\ }  \\
        \cline{2-3}
         & Babel  & \makecell[l]{ Amharic (am),\\ Cantonese (yue),\\Georgian (ka), Guarani (gn),\\ Kurmanji-kurdish (ku),\\ Mongolian (mn), Pashto (ps), \\Swahili (sw), Tagalog (tl) } \\
         \midrule
         \multirow{5}{*}{Distillation} & MATERIAL & Farsi (fa), Somali (so) \\ 
         \cline{2-3}
         & Babel & \makecell[l]{Amharic (am), Georgian (ka), \\Guarani (gn), Javanese (jv), \\Kazakh (kk), Mongolian (mn), \\ Pashto (ps), Vietnamese (vi)} \\ 
        \bottomrule
    \end{tabular}
    \label{tab:data}
    \vspace{-2em}
\end{table}

\textbf{Splicing Setup}. The GMM-HMM used to generate the timestamps is trained using kaldi's Librispeech recipe\footnote{https://github.com/kaldi-asr/kaldi/blob/master/egs/librispeech/s5/run.sh} and we used the tri6b model for alignment. The syllable boundaries are generated from a syllabified CMU dictionary\footnote{http://webdocs.cs.ualberta.ca/~kondrak/cmudict.html}. During training, 37.5\% of the utterances are randomly shuffled.

\textbf{Distillation Setup}. The proposed DistilXLSR model consists of a 6-layer CNN feature extractor and a 12-layer transformer encoder. The transformer encoder is initialized by layer-jumping initialization according to Eq. \ref{distil}. We also apply the masked speech denoising strategy according to WavLM where 15\% of the utterances are mixed with another one in the same batch \cite{chen_wavlm_2022}. The distillation is performed on an RTX 3090 GPU for 200k updates and around 37 hours with a batch size of 6 utterances and a learning rate of 2.0e-4.

\textbf{Fine-tuning Setup}. We fine-tune the models using the Fairseq toolkit following the experimental settings of Zhao et.al \cite{zhao_improving_2022}. For each language, we add a linear layer on the top and optimize the model using the Connectionist Temporal Classification (CTC) loss. Parameters are updated every 8 steps and the model is trained for 20k updates and around 5 hours. The learning rate is set to 1.0e-4 with a tri-stage rate schedule, where the learning rate increases linearly to the set value for the first 2k updates, holds constant for the next 8k updates, and decreases linearly to 0 for the remaining updates. The batch size is set to 1.28M samples, while 55\% of the frames and 25\% of the channels of the CNN features are masked.

\section{Results}

\subsection{Comparing With Teacher Models}

Table \ref{main result} shows the performance of our proposed model on 15 low-resource languages. Using the teacher models as benchmarks, we do not find significant gaps in the performance across languages and teacher models, demonstrating the generalizability of our approach. Our proposed models achieve lower word error rates on the Common Voice dataset, and the degradation compared to the teacher models is relatively small. In an extremely low-resource setting, where only a 5-hour training set is available for each Common Voice language, the WER of the proposed model is only 2.5\% higher than the XLSR-53 teacher model, while the average error rate increases by 4.18\% in absolute terms.

The degradation is more obvious in the Babel and MATERIAL datasets. As mentioned above, the Babel and MATERIAL datasets consist mainly of telephone conversations at a sample rate of 8KHz, and the signal-to-noise ratio (SNR) is much higher than that of the Common Voice dataset. Due to the lower complexity and fewer parameters, the compressed models are more prone to underfitting and more sensitive to the noise in the data. Although we have applied some data augmentation strategies such as masked speech denoising, the degradation is still present. For the XLS-R teacher, on the 10 languages of the Babel and MATERIAL datasets, the average WER is 47.06\%, which is 5.3\% lower than the student model. 

\begin{figure}
    \centering
    \includegraphics[width=0.8\linewidth]{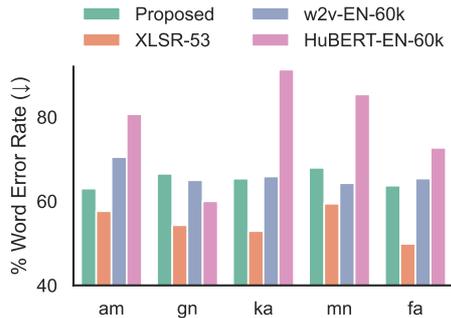}
    \vspace{-2em}
    \caption{Word error rates of 5 Babel languages of the HuBERT-EN-60k \cite{hsu_hubert_2021}, w2v-EN-60k \cite{baevski_wav2vec_2020}, XLSR-53, and the proposed method inferred without language models. Results of the first 2 models are from Zhao et al. \cite{zhao_improving_2022}.}
    \label{scatter}
    \vspace{-2em}
\end{figure}

Despite this, our model still retains the cross-lingual representation ability. Figure \ref{scatter} provides a visualization of the WERs on the Babel and MATERIAL datasets of 4 different pre-trained models. The results show that our model can achieve comparable or even better performance than the w2v-EN-60k and HuBERT-EN-60k models, which are pre-trained on a 60,000-hour English dataset.
The experiments on 15 languages demonstrate the cross-lingual representation ability of the proposed models, even though they are trained on 960h of English data. Our model requires only 1 GPU for training with 50\% fewer parameters, demonstrating the trade-off between training cost, computation, and performance. \footnote{Our preliminary experiments also shows that DistilXLSR significantly outperforms E2E or Hybrid Models with same amount of labeled data. Detailed results can be found at our github page.}

\begin{table*}[!ht]
    \centering
    \caption{The word error rate for 15 low-resource languages. Fine-tuning parameters are set according to the best results in OpenASR21. S1 and S2 are distilled from XLSR-53 and XLS-R, respectively.}
    \resizebox{\textwidth}{!}{
    \begin{tabular}{lcccccccccccccccc}
    \toprule
        \multirow{2}{*}{Model} & \multicolumn{15}{c}{Languages} & \multirow{2}{*}{Avg.} \\ \cline{2-16}
        & el & nl & eu & ia & pl & ta & ps & ku & sw & tl & am & gn & ka & mn & fa &  \\ 
        \midrule
        XLSR-53 & 10.7 & 12.4 & 29.5 & 27.1 & 25.5 & 65.5 & 45.5 & 65.1 & 40.5 & 43.9 & 47.7 & 41.2 & 41 & 46.4 & 33.8 & 38.38\\ \hline
        S1 & 14.2 & 14.9 & 33.8 & 34.4 & 28.8 & 69.8 & 50.5 & 65.6 & 45.3 & 49.8 & 50.6 & 48.5 & 47.7 & 52.8 & 43 & 43.31\\ \hline
        XLS-R & 9.0 & 13.4 & 28.2 & 25.2 & 24.7 & 63 & 43.1 & 61.2 & 37.2 & 41.1 & 41.4 & 38.9 & 38.4 & 43.3 & 32.6 & 36.04 \\ \hline
        S2 & 13.2 & 14.6 & 29.4 & 34.8 & 28.9 & 67.7 & 49.2 & 67.2 & 43.8 & 48.2 & 48.6 & 46.2 & 45.7 & 50.6 & 40.6 & 41.91 \\ 
    \bottomrule
    \label{main result}
    \end{tabular}
    }
    \vspace{-1.5em}
\end{table*}

\subsection{Ablation Studies}
\subsubsection{The Effectiveness of Data Splicing\label{section:splicing}}
\begin{figure}
	\centering
        \subfloat[The effectiveness of data splicing.\label{scatter1}]{\includegraphics[width=0.9\linewidth]{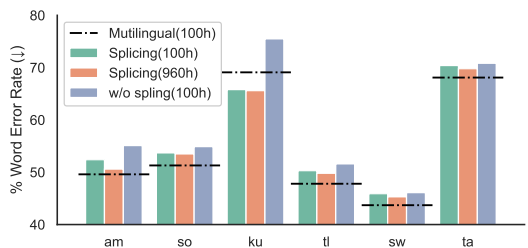}} \\
        \subfloat[The effectiveness of layer-jumping initialization.\label{scatter2}]{\includegraphics[width=0.9\linewidth]{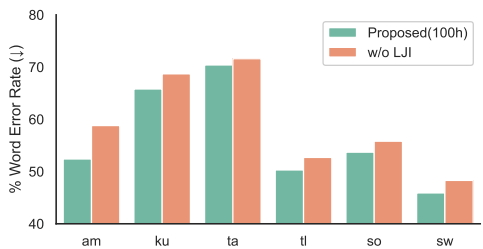}}
        \caption{WERs for ablation studies, inferred with 4-gram LMs.} 
        \vspace{-1.5em}
 \end{figure}

Figure \ref{scatter1} compares the performance of 4 models distilled from the XLSR-53 teacher. For all 6 low-resource languages, the application of data splicing reduces the word error rates, especially for the Kurmanji-Kurdish, where models with data splicing outperform the multilingual distillation model while the model without data splicing does not. This phenomenon demonstrates the effectiveness of data splicing. Moreover, using a larger amount of data can bring further improvement, suggesting the possibility of using large-scale unsupervised English datasets, such as the Libri-light \cite{kahn_libri-light_2020} or Gigaspeech \cite{chen_gigaspeech_2021}, to improve distillation or even pre-training.

For the multilingual distillation model, Amharic (am) and Somali (so) appear in the training set while Swahili (sw) and Tamil (ta) do not. Kurmanji-Kurdish (ku) and Tagalog (tl) are also absent, but both of them have similar languages from the same language family in the training set (Pashto and Farsi for Kurmanji-Kurdish, and Vietnamese for Tagalog). However, we do not find that the models behave differently in these 6 languages, which shows that the multilingual distillation model, and the proposed data splicing model, have learned the language-independent cross-lingual representation ability from the XLSR-53 teacher.

\subsubsection{The effectiveness of Layer-Jumping Initialization}

\begin{figure}[ht]
	\label{cka}
	\centering
        \subfloat[w2v2-base\label{base}]{\includegraphics[width=.42\linewidth]{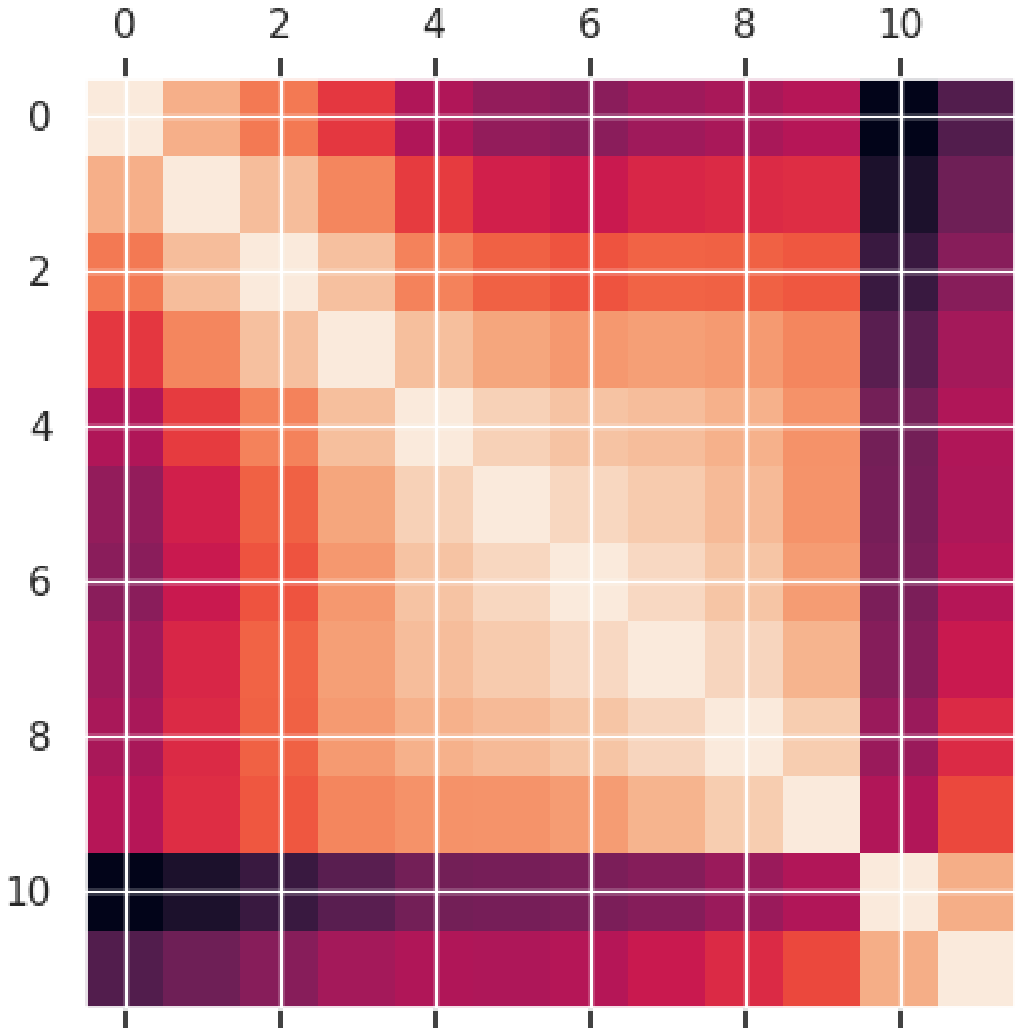}}\hspace{5pt}
        \subfloat[XLSR-53\label{xlsr}]{\includegraphics[width=.42\linewidth]{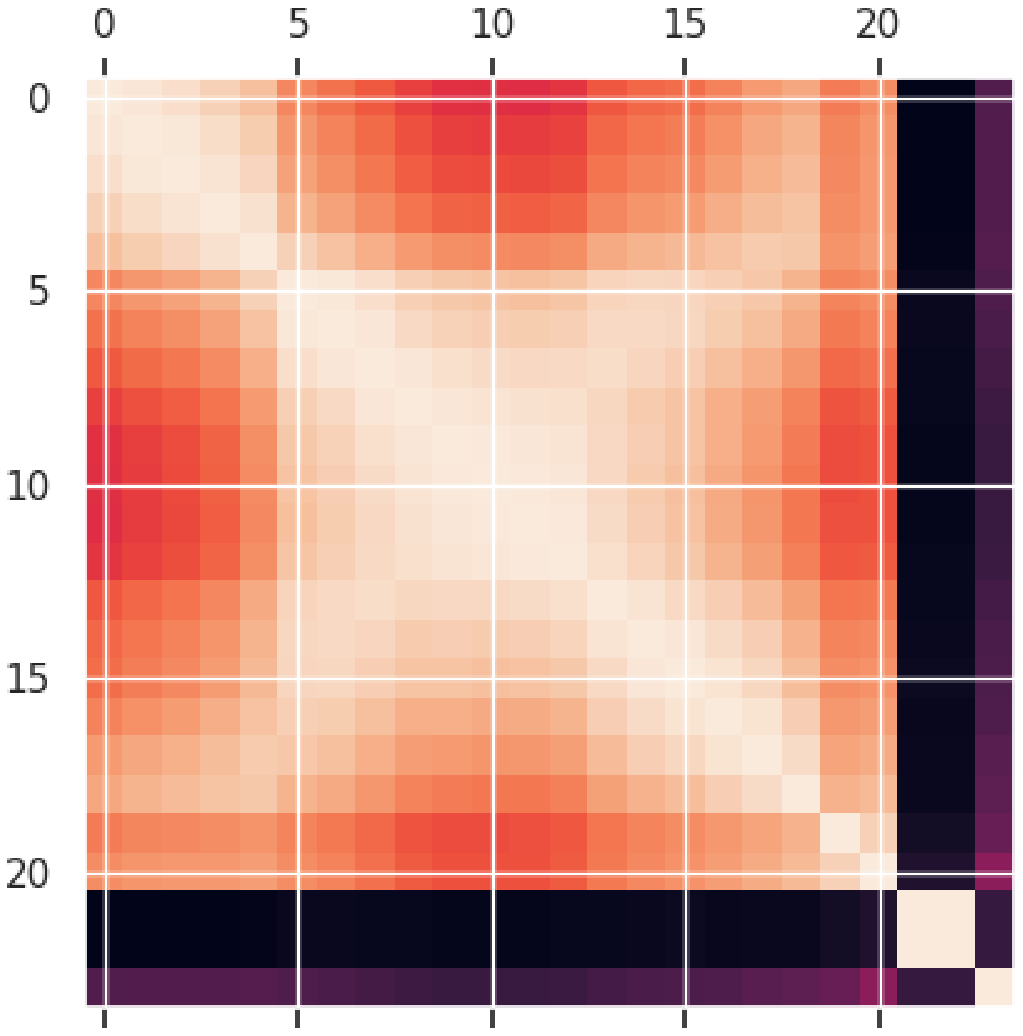}} \\
        \subfloat[w/ LJI\label{wl}]{\includegraphics[width=.42\linewidth]{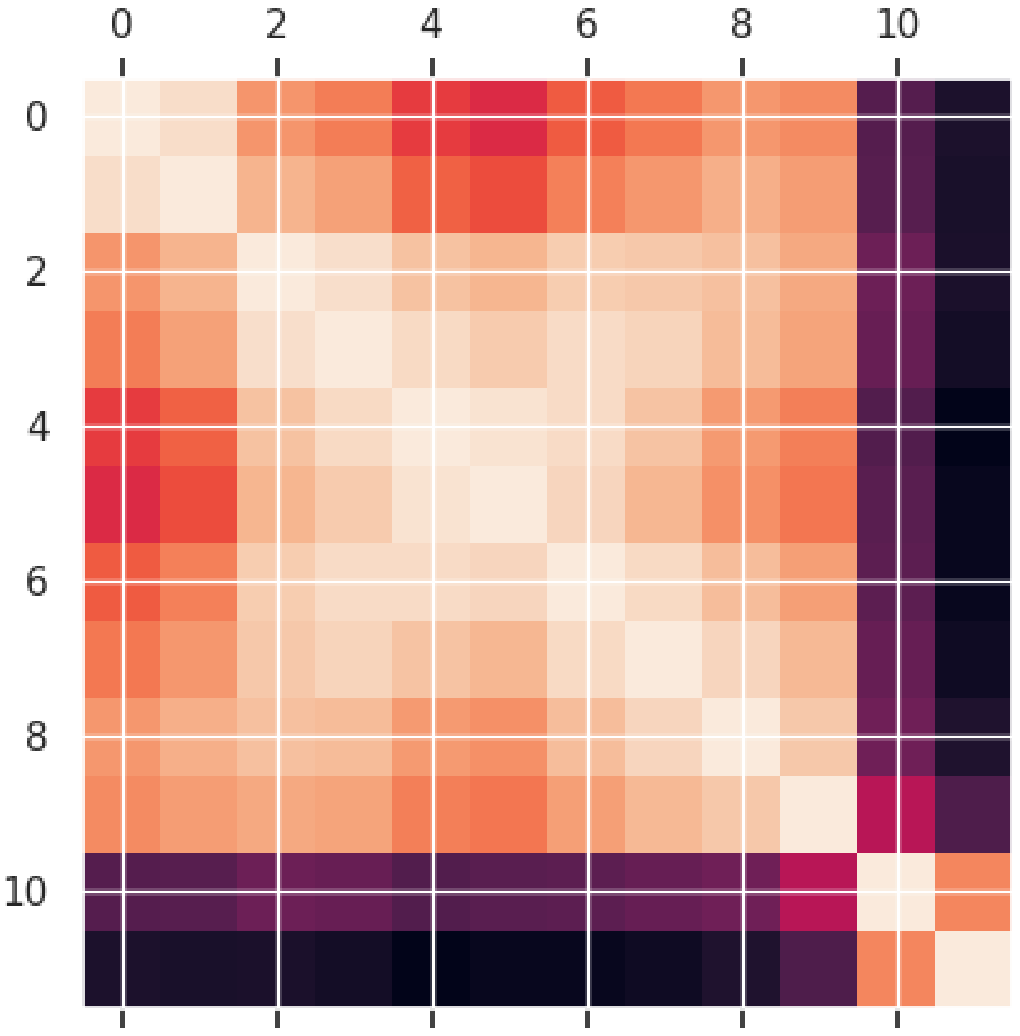}} \hspace{5pt}
        \subfloat[w/o LJI\label{wol}]{\includegraphics[width=.42\linewidth]{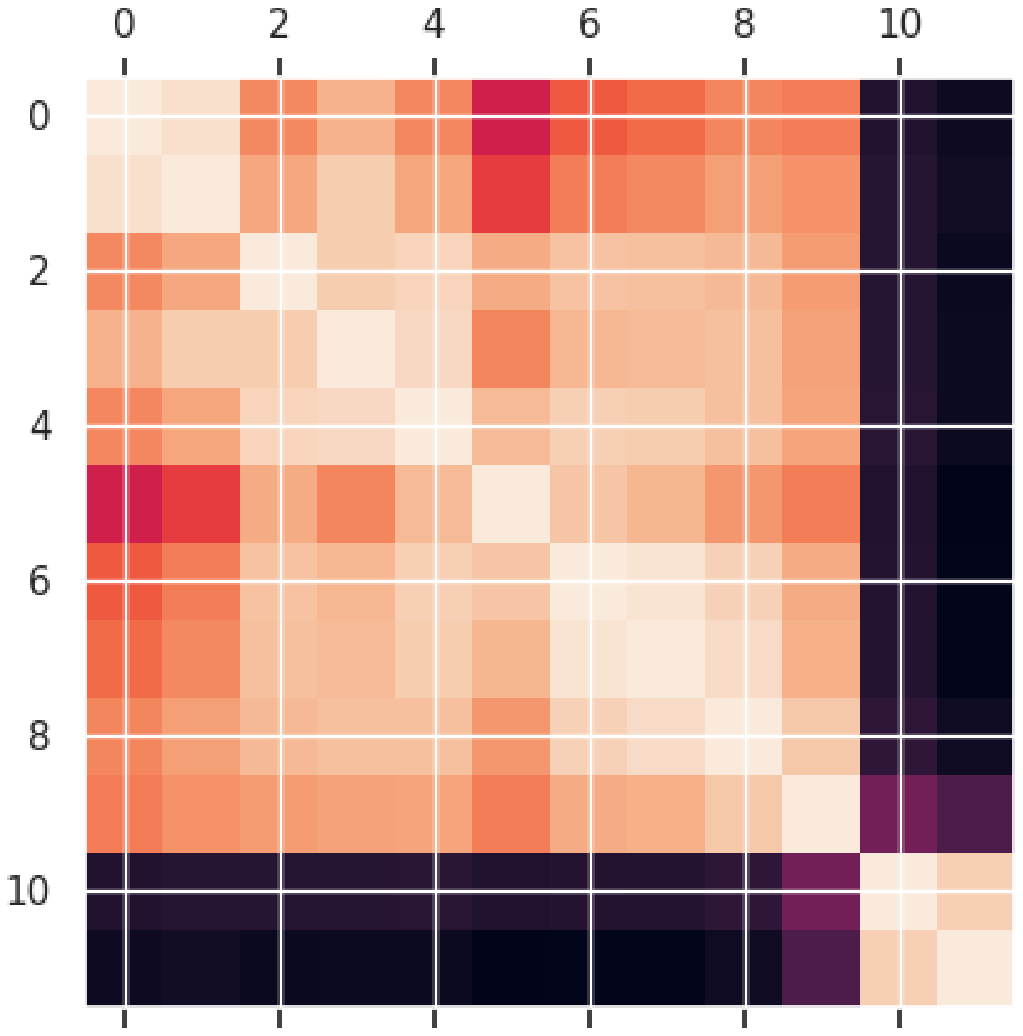}}
        \caption{CKA interlayer similarities. LJI is for Layer-Jumping Initialization. LJI allows the model to better captures the interlayer similarity of the teacher model.} 
        \vspace{-1.5em}
 \end{figure}

 Figure \ref{scatter2} compares the performance of the proposed model with continuous (e.g. 0-11) initialization and layer-jumping initialization, where we can observe a significant increase in the WERs for all the languages, proving the importance of fully exploiting the pre-trained weights. Figures \ref{wl} and \ref{wol} show the CKA interlayer similarities of these two models. It can be seen that the model with layer-jumping initialization better captures the interlayer similarity of the teacher model. In addition, the layer-jumping initialization allows the proposed model to learn the differences between the 22nd/23rd and the 24th layers, which is unclear without the layer-jumping initialization.
 
\section{Discussions}
The performance degradation on the Babel and MATERIAL datasets illustrates the importance of solving the underfitting problem. Structured pruning, although not yet successfully applied to large-scale pre-trained acoustic models, may have the potential to further preserve the performance without dedicated hardware. In addition, it is useful to validate the effectiveness of data splicing on large English datasets. We leave these questions for future work.

\section{Conclusions}
In this paper, we propose a method to distill cross-lingual speech representation model using only English data. Our experiments on 15 low-resource languages show that our proposed model can maintain the cross-lingual representation ability with 50\% fewer parameters. Further experiments demonstrate the effectiveness of using layer-jumping initialization and applying data splicing. Our method provides compressed cross-lingual representation models and is also able to improve the cross-lingual performance of the English pre-trained models.

\bibliographystyle{IEEEtran}
\bibliography{citations_mld}

\end{document}